\documentclass[letterpaper, 10 pt, conference]{ieeeconf}  

\IEEEoverridecommandlockouts                              

\usepackage{cite}
\usepackage{amsmath} 
\usepackage{amssymb}  
\usepackage{CJKutf8}
\usepackage{pgfplots} 
\usepackage{algorithm}
\usepackage{booktabs}
\usepackage{algpseudocode}
\usepackage{hyperref} 
\usepackage[font=footnotesize,labelsep=colon]{caption}
\title{\LARGE \bf
RoboSeek: You Need to Interact with Your Objects
}

\author{%
  \parbox{\linewidth}{\centering
    {\large Yibo Peng$^{1,2,5*}$, Jiahao Yang$^{1*}$, Shenhao Yan$^{1,3}$, Ziyu Huang$^{1,2}$, Shuang Li$^{5}$, Shuguang Cui$^{1,2}$,}\\
    {\large Yiming Zhao$^{1,4,5\dagger}$, Yatong Han$^{1,5\dagger}$}\\[6pt]
    \normalsize
    $^{1}$ FNii-Shenzhen \quad $^{2}$ The Chinese University of Hong Kong, Shenzhen \quad $^{3}$ Northeastern University\\
    $^{4}$ Harbin Engineering University \quad $^{5}$ Infused Synapse AI\\[6pt]
    \texttt{yiming\_zhao@hrbeu.edu.cn \quad hanyatong@cuhk.edu.cn}
  }%
  \thanks{* Equal contribution. \quad $\dagger$ Corresponding authors.}%
}

\begin{document}
\begin{CJK*}{UTF8}{gbsn} 

\maketitle
\thispagestyle{empty}
\pagestyle{empty}

\begin{abstract}
Optimizing and refining action execution through exploration and interaction is a promising way for robotic manipulation. However, practical approaches to interaction-driven robotic learning are still underexplored, particularly for long-horizon tasks where sequential decision-making, physical constraints, and perceptual uncertainties pose significant challenges. Motivated by embodied cognition theory, we propose \textbf{RoboSeek}, a framework for embodied action execution that leverages interactive experience to accomplish manipulation tasks. RoboSeek optimizes prior knowledge from high-level perception models through closed-loop training in simulation and achieves robust real-world execution via a real2sim2real transfer pipeline. Specifically, we first replicate real-world environments in simulation using 3D reconstruction to provide visually and physically consistent environments., then we train policies in simulation using reinforcement learning and the cross-entropy method leveraging visual priors. The learned policies are subsequently deployed on real robotic platforms for execution. RoboSeek is hardware-agnostic and is evaluated on multiple robotic platforms across eight long-horizon manipulation tasks involving sequential interactions, tool use, and object handling. Our approach achieves an average success rate of 79\%, significantly outperforming baselines whose success rates remain below 50\%, highlighting its generalization and robustness across tasks and platforms. Experimental results validate the effectiveness of our training framework in complex, dynamic real-world settings and demonstrate the stability of the proposed real2sim2real transfer mechanism, paving the way for more generalizable embodied robotic learning. \textbf{Project Page:} \href{https://russderrick.github.io/Roboseek/}{https://russderrick.github.io/Roboseek/}

\end{abstract}

\section{Introduction}

\begin{figure}[t]
    \centering
    \includegraphics[width=0.5\textwidth]{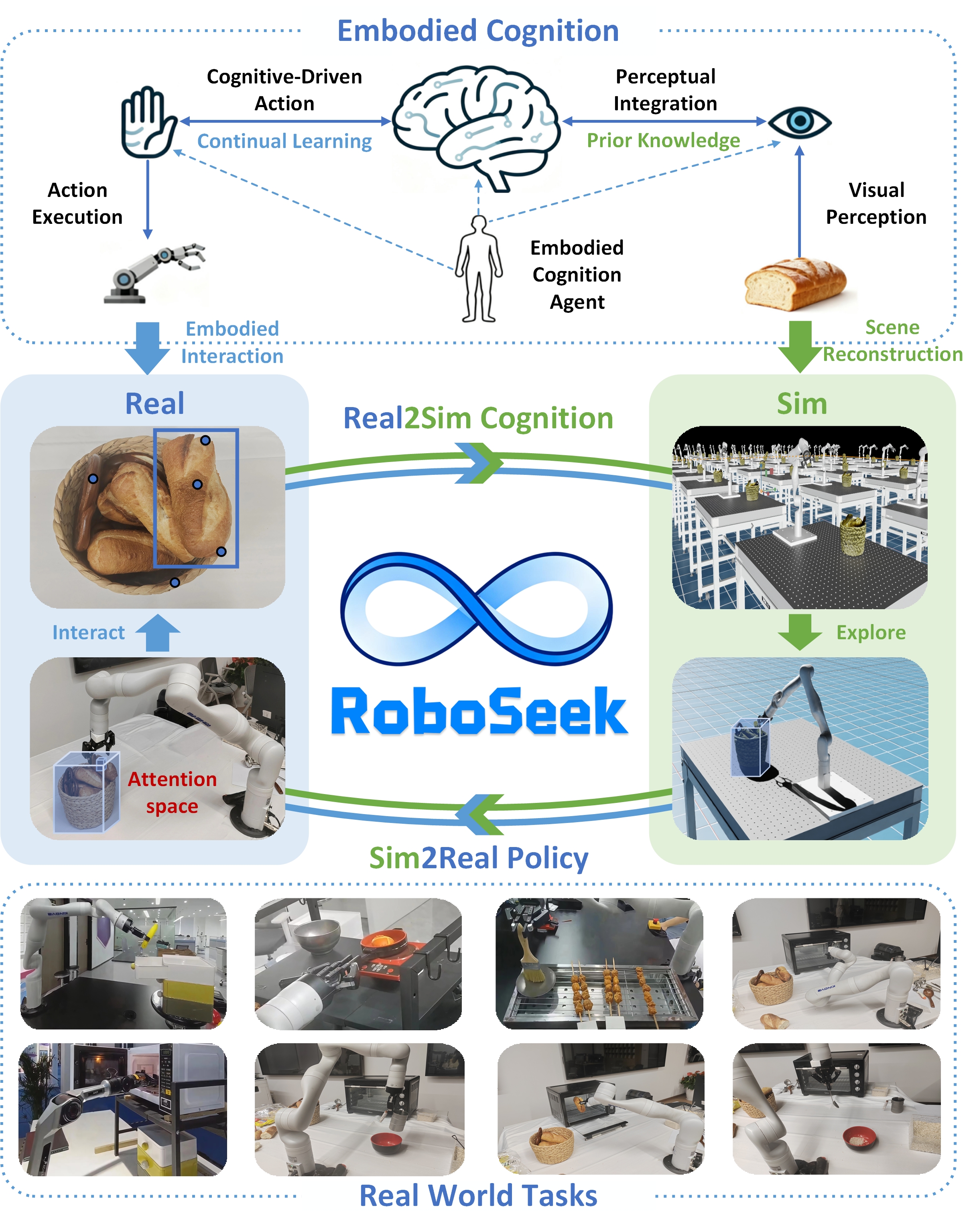}
    \caption{\textbf{Roboseek}: an interaction-driven framework inspired by embodied cognition. The system is capable of closed-loop control and is designed to perform a variety of long-horizon, real-world tasks with stable and robust action execution.}
    \label{fig:overview}
\end{figure}
Enabling generalizable robotic manipulation in open-world settings is a central challenge in embodied intelligence \cite{liu2025aligning}. This goal is deeply rooted in embodied cognition theory \cite{clark1998being,foglia2013embodied}, which posits that cognitive processes such as task understanding and action planning are not isolated from the physical body and environment but emerge from dynamic interactions between an agent’s body, objects, and surrounding context \cite{shridhar2020alfred, li2023behavior}. Unlike abstract symbolic reasoning, embodied cognition emphasizes that an agent’s ability to perceive object functionalities (e.g., a drawer’s openability or a spoon’s scooping potential) and execute goal-directed actions depends on real-time sensory feedback from physical interaction. This principle is particularly critical for long-horizon, dexterous manipulation tasks. As shown in Figure~\ref{fig:overview}, such tasks include opening a drawer to place an object inside and then closing it, or scooping cereal, pouring it into a bowl, and adding milk. These tasks require not only tool use and object transport but also the exploitation of affordance \cite{gibson2014theory}. Affordance is a task-specific object property detectable only through embodied engagement, and affordances include a drawer handle for opening or a spoon handle for scooping. Consequently, a successful robotic agent must integrate visual perception, task-level planning, and low-level motor execution in a way that aligns with the interactive, context-dependent nature of embodied cognition \cite{haigh1996using,masterman2024landscape}.

Large language models (LLMs) and vision language models (VLMs) have recently advanced significantly, greatly improving agents’ capabilities in visual understanding and language-conditioned reasoning \cite{achiam2023gpt,driess2023palm,bai2025qwen2,chen2023towards}. However, these models often operate on static datasets and lack mechanisms to ground their reasoning in embodied interaction. This constitutes a fundamental limitation from the perspective of embodied cognition. End-to-end vision language action models (VLAs) have shown promise in bridging perception and action \cite{kim2024openvla,zitkovich2023rt,black2024pi0}, but they still face challenges in language following and stable control because they cannot leverage continuous feedback from physical interaction to refine actions. Other methods that combine VLM-based planners \cite{liu2024moka,huang2024rekep,liu2025robodexvlm} with separate motion solvers fare no better: they rely heavily on precomputed task interpretations and static object representations while ignoring the embodied cognition principle that object affordances and action strategies must be updated through real-time interaction.

Guided by embodied cognition theory, we argue that the complexity of the physical world necessitates continual learning through active exploration and interaction \cite{wang2024comprehensive}. For robots, this means abandoning static, precomputed representations of objects or tasks. Instead, they must dynamically construct and iteratively refine task-relevant visual representations on the fly via physical engagement with objects and observing the resulting changes in the environment. Toward this goal, we introduce the concept of an attention space—a dynamic, interaction-driven visual representation that encodes potential task-specific affordances for objects within the workspace (e.g., the handle of a drawer). This attention space directly embodies the core of embodied cognition: it is not a fixed model of the world but a flexible construct that evolves as the robot interacts with objects and receives environmental feedback.

Reinforcement learning (RL) \cite{sutton1998reinforcement} aligns naturally with this embodied paradigm, as it optimizes policies through trial-and-error feedback from the environment. This process mirrors how embodied agents learn to adapt actions to context. Building on previous work \cite{patel2025real,huang2024rekep,liu2024moka}, we introduce RoboSeek, an embodied action execution module explicitly designed to operationalize the principles of embodied cognition in robotic manipulation. RoboSeek performs three key functions. First, it constructs an initial attention representation for visual perception based on keypoints. This construction is grounded in early sensory input, in line with embodied cognition’s emphasis on sensory-motor coupling. Second, it trains an embodied executor via RL to explore and act within this attention space, with the executor learning action policies through physical interaction. Third, it iteratively refines the attention space using the cross-entropy method (CEM) \cite{rubinstein1999cross}, and this refinement is guided by the executor’s interaction feedback to update representations as the robot gains embodied experience. The optimized attention space and executor are then deployed on real-world hardware to perform manipulation tasks.

Given the rapid development of robotic simulators and the risks and instability of real-world online training \cite{todorov2012mujoco,mittal2023orbit}, we adopt a real2sim2real pipeline. This pipeline is a practical implementation of the "simulated rehearsal for real interaction" logic in embodied cognition. We bootstrap and validate our approach in simulation, where simulation enables safe and scalable exploration to support this process. We then apply sim2real transfer techniques to bridge the gap between simulated and real embodied experiences before deploying the learned models on physical platforms. We summarize a set of implementation strategies that substantially reduce the sim2real gap, and these strategies are validated through extensive experiments. Our approach demonstrates reliable and robust performance across multiple robotic platforms and a variety of long-horizon manipulation tasks, thereby providing empirical evidence for the value of embedding embodied cognition principles in robotic systems.

Our contributions are summarized as follows:

\begin{enumerate}
    \item We propose an attention space formulation for embodied execution, which enables the learning of keypoint-based visual priors and their subsequent refinement through a task-driven closed-loop process.
    \item We develop a RL-based embodied executor, integrated with a CEM-driven refinement loop, to realize the joint optimization of the attention space and the executor’s control policies.
    \item We design a hardware-agnostic real2sim2real training and transfer pipeline, which achieves robust sim-to-real performance across multiple robotic platforms and long-horizon manipulation tasks, and supports safe and scalable learning for embodied systems before real-world deployment.
\end{enumerate}
 
\section{Related Work}
\subsection{Foundation models for robotics}
Foundation models have shown great promise for robotics by enabling generalizable behaviors through large-scale pretraining \cite{achiam2023gpt,team2023gemini,bai2025qwen2,ji2025robobrain,zitkovich2023rt, kim2024openvla,black2024pi0}. Building on the successes of LLMs and VLMs, recent work has extended these paradigms toward VLAs that aim to produce general-purpose, embodied agents capable of high-level perception and task planning. However, despite strong advances in spatial perception and reasoning, the low-level action execution capabilities remain limited. Moreover, existing foundation models are typically trained on static datasets and therefore lack mechanisms for continuously learning from online environment interactions to refine both visual understanding and action control. To address this gap, our method enhances foundation models with a low-level action module that learns from exploration and interaction. Using the feedback information from the environment, we provide foundation models with a closed-loop error correction mechanism and stable action execution capability.
\begin{figure*}[t]
    \centering
    \includegraphics[width=1\textwidth]{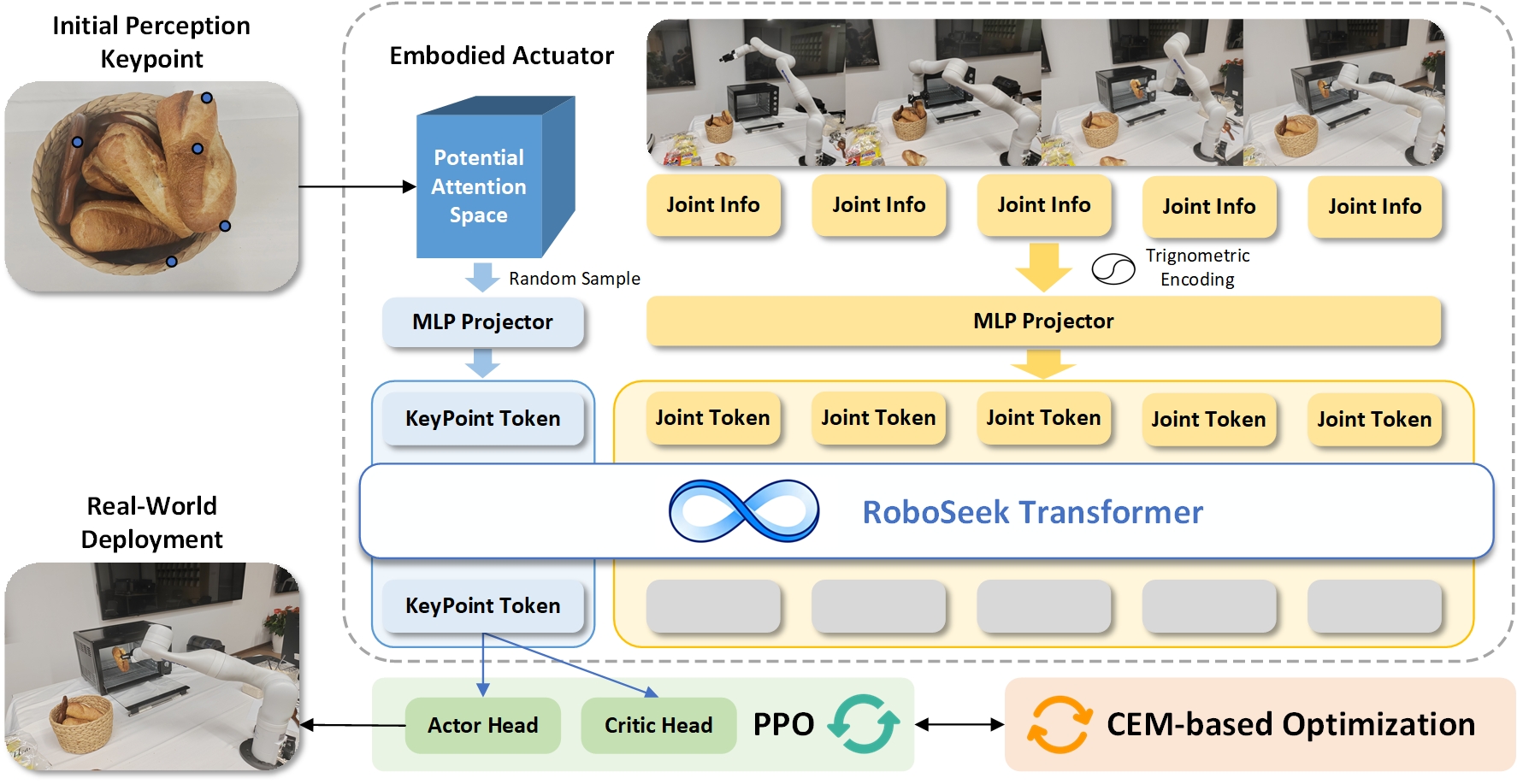}
    \caption{\textbf{Overview of Roboseek's framework}: Roboseek is a closed-loop real2sim2real pipeline that leverage visual priors to performe stable action through interaction-driven methods.}
    \label{fig2}
\end{figure*}
\subsection{Keypoint-based affordance}
Affordance means the potential actions that an object offers to an agent, based on the object’s properties and the agent’s capabilities \cite{gibson2014theory}. In robotic manipulation, affordance learning refers to the process of identifying task-specific semantic regions of an object that enable effective interaction. Keypoints are commonly adopted as compact and structured representations for encoding affordance information. Prior approaches have demonstrated the utility of keypoints for capturing semantic affordances, yet they exhibit several notable limitations. Some methods rely on planning-based action \cite{liu2024moka,huang2024rekep,liu2025robodexvlm,pan2025omnimanip,liu2025kuda}, which constrains adaptability in dynamic environments; others require extensive data collection, resulting in high annotation and computational costs \cite{yuan2024robopoint,fang2024keypoint,wang2025skil,ma2025glover++}; while many fail to incorporate closed-loop mechanisms, limiting their robustness during interaction \cite{kuang2024ram,patel2025real,xu2025a0,su2025resem3d}. In contrast, our work introduces an automatic closed-loop learning framework that addresses these challenges, providing a new perspective on keypoint-based affordance learning.
\subsection{RL for robotic manipulation}
RL has achieved remarkable progress in robotic manipulation, enabling agents to acquire control policies through interaction with the environment. Despite these advances, several limitations persist in prior works. Many approaches rely on offline RL \cite{lee2024affordance,kumar2022pre,chen2025conrft}, which requires collecting and annotating large datasets, thereby incurring substantial data and labeling costs. Others focus primarily on training in simulation \cite{lu2025vla,shu2025rftf,tan2025interactive}, often leading to poor transferability to real-world robots. Furthermore, a number of studies employ simple network architectures and naive algorithms \cite{patel2025real, khazatsky2021can}, restricting their applicability to relatively basic tasks such as pick-and-place. Building on these insights, we develop a hardware-free RL framework that learns directly through exploration and interaction, and enables robust sim2real transfer.

\section{Method}

In this section, we formally propose RoboSeek, an approach that leverages RL and CEM to learn the attention space and execute low-level actions for manipulation tasks with automatically closed-loop recorrection. We sequentially introduce our RL-driven embodied actuator in the attention space and the attention space optimizer by CEM. Then, we discuss our hardware-free real2sim2real transfer mechanism. Finally, we demonstrate how our method works in long-horizon manipulation tasks.  Our
 method overview is illustrated in Figure~\ref{fig2}.

\subsection{Problem Statement}
Grounded in the theory of embodied cognition, we view manipulation as a process that inherently couples perception, bodily action, and interaction with the environment. In this perspective, we employ semantic keypoints as unified representations of affordances that emerge from the agent’s embodied interaction with the world. These keypoints act as an interface between high-level perceptual inference and low-level embodied action execution.  

Formally, we define the attention space as a 3D workspace that contains all candidate semantic keypoints:
\[
\mathcal{A} \subset \mathbb{R}^3,
\]
where each semantic keypoint is denoted by 
\[
k \in \mathcal{A}.
\]

We cast long-horizon manipulation as a two-stage problem: (1) identify semantic keypoints $\{k_t\}_{t=1}^T$ for each step, and (2) execute actions $\{a_t\}_{t=1}^T$ conditioned on these keypoints. The policy of the embodied actuator is represented as
\[
a_t = \pi_\theta(s_t, k_t),
\]
where $s_t$ is the state at time $t$ and $\pi_\theta$ is a parameterized policy.  

Our objective consists of two parts. First, we train an embodied actuator $\pi_\theta$ that is capable of exploring freely in the attention space $\mathcal{A}$:
\[
a_t \sim \pi_\theta(s_t, k), \quad k \in \mathcal{A}.
\]
Second, we optimize the distribution $p(k)$ of the attention space based on environmental feedback, so that it converges toward the optimal keypoint $k^*$:
\[
k^* = \arg\max_{k \in \mathcal{A}} \; \mathbb{E}\Big[ R(s_{t+1} \mid s_t, \pi_\theta(s_t, k)) \Big].
\]

\subsection{RL-driven embodied actuator in attention space}
We use high-level perception models to give an initial prediction of keypoint. The goal of the embodied actuator is to be able to move to any area within a uniform distribution using the initial keypoint as the mean.

We train our actuator in a simulation environment with RL. We use IsaacLab \cite{mittal2023orbit} as the learning platform, with the Proximal Policy Optimization (PPO) algorithm \cite{schulman2017proximal} employed as the training framework and the transformer architecture \cite{vaswani2017attention} used as the policy network. Our experiments show that the self-attention mechanism of transformer can well learn the cooperative movement between different joints of the manipulator and understand the pose difference between joints and keypoints. The network input is a concatenation of the trigonometric function encoding of the manipulator's current joint angles, historical actions, and randomly sampled poses within the distribution space. The use of trigonometric function encoding can avoid the periodic ambiguity of angles and capture more relative position information. We treat the sampled keypoint pose as the first token, followed by each joint angle in sequence. The pose of the keypoint is encoded into its embedding information via a lightweight encoder. Similarly, the angles of each joint and previous actions are encoded into vector embeddings of a specified dimension as joint information. A six-layer transformer with three attention heads is used to learn the pose differences between joints and the keypoint. After the keypoint token output by the transformer, two Multi-Layer Perceptron (MLP) action heads are added to output the Actor and the Critic of the PPO algorithm respectively. The Actor outputs actions as the relative angles of each joint's movement per step. We use Exponential Linear Unit (ELU) \cite{clevert2015fast} as the activation function.

We carefully design a scalar reward function combining the following terms with different weights:

\noindent\textbf{Gripper-Keypoint Distance Reward.}  
Let $\mathbf{p}_t^{\text{ee}}\in\mathbb{R}^3$ denote the end-effector position at time step $t$, and $\mathbf{p}_t^{\star}\in\mathbb{R}^3$ the target keypoint position.  
The Euclidean distance is
\[
    d_t \;=\; \left\lVert \mathbf{p}_t^{\text{ee}} - \mathbf{p}_t^{\star} \right\rVert_2 .
\]
To provide both coarse guidance and fine precision shaping, we also employ two $\tanh$-kernels with different scales:
\[
    r_t^{\text{dist}} \;=\; 
    w_{d1}{d_t} \;+\;
    w_{d2}\!\left(1-\tanh\!\left(\tfrac{d_t}{\sigma_1}\right)\right) \;+\;
    w_{d3}\!\left(1-\tanh\!\left(\tfrac{d_t}{\sigma_2}\right)\right),
\]
where $(\sigma_1,\sigma_2)=(0.3,\,0.05)$.

\noindent\textbf{Gripper-Keypoint Orientation Reward.}  
Let $q_t^{\text{ee}}\in\mathbb{S}^3$ be the end-effector quaternion and $q_t^{\star}\in\mathbb{S}^3$ the commanded orientation.  
We measure the geodesic distance on $\mathbb{S}^3$:
\[
    \theta_t \;=\; 2\,\arccos\!\big(\,|\langle q_t^{\text{ee}},\, q_t^{\star}\rangle|\,\big),
\]
and define the orientation reward as
\[
    r_t^{\text{ori}} \;=\; -\,w_{\text{ori}}\,\theta_t , 
\]

\noindent\textbf{Joint Action Reward.}  
To penalize excessive actuation and encourage smooth control, we regularize action magnitude, action rate, and joint velocity.  
Let $\mathbf{a}_t$ denote the action and $\dot{\mathbf{q}}_t$ the joint velocities.  
The regularizer is
\[
    r_t^{\text{act}} \;=\; 
    -\,w_{\ell2}\,\|\mathbf{a}_t\|_2^2 
    \;-\; w_{\text{rate}}\,\|\mathbf{a}_t-\mathbf{a}_{t-1}\|_2^2
    \;-\; w_{\text{vel}}\,\|\dot{\mathbf{q}}_t\|_2^2,
\]

The final scalar reward is the weighted sum of all components:
\[
    r_t \;=\; r_t^{\text{dist}} \;+\; r_t^{\text{ori}} \;+\; r_t^{\text{act}} .
\]

Since dense rewards may lead the policy to fall into local optima and converge early, we adopt a curriculum learning strategy that dynamically adjusts the weights of different reward terms over training steps \cite{florensa2017reverse}. This design encourages a more balanced trade-off between exploration and convergence, ultimately resulting in more stable motion control.
\subsection{Attention space optimizer with CEM}

\begin{figure}[t]
    \centering
    \includegraphics[width=0.5\textwidth]{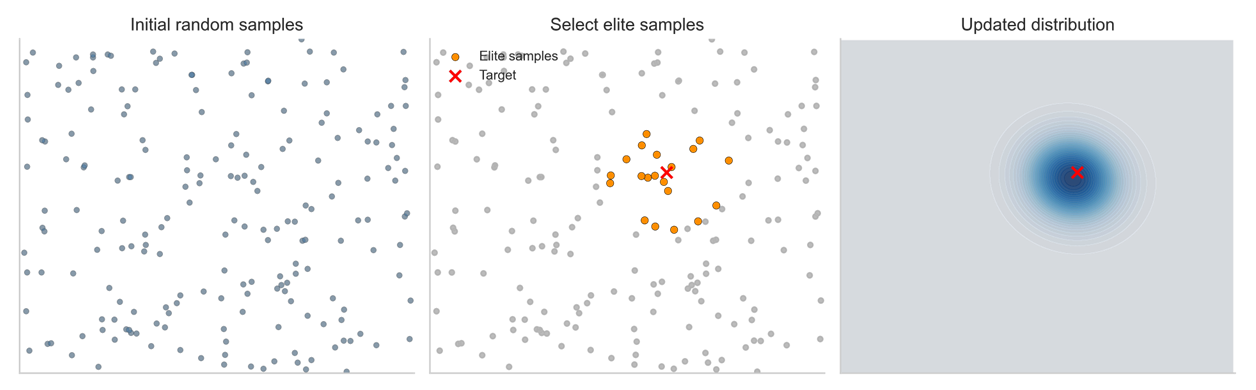}
    \caption{\textbf{Visualization of CEM.} The Cross-Entropy Method is based on Monte Carlo sampling, and iteratively updates and converges the distribution of the attention space under the guidance of the reward function. }
    \label{CEM}
\end{figure}

After training the embodied actuator, we refine and optimize the attention space using the Cross-Entropy Method \cite{rubinstein1999cross}, a sampling-based optimization algorithm. Visualization of CEM is shown in Figure~\ref{CEM}.

We model the attention space as a Gaussian distribution
\[
x \sim \mathcal{N}(\mu_t, \Sigma_t), \quad x \in \mathbb{R}^d,
\]
where $\mu_t$ and $\Sigma_t$ denote the mean and covariance at iteration $t$.  

At each iteration, we first sample $m$ candidate keypoints from the current distribution:
\[
\mathcal{X}_t = \{x_i\}_{i=1}^m, \quad x_i \sim \mathcal{N}(\mu_t, \Sigma_t).
\]

For each candidate $x_i$, the actuator executes $n$ rollouts under the task-specific reward function $r(\cdot)$, and the average return is computed as
\[
R(x_i) = \frac{1}{n} \sum_{j=1}^n r(x_i, j).
\]

We then select the top-$k$ candidates with the highest average returns, denoted by the index set $\mathcal{S}_t$, and update the Gaussian distribution accordingly:
\[
\mu_{t+1} = \frac{1}{k} \sum_{i \in \mathcal{S}_t} x_i,
\]
\[
\Sigma_{t+1} = \frac{1}{k} \sum_{i \in \mathcal{S}_t} (x_i - \mu_{t+1})(x_i - \mu_{t+1})^\top.
\]

This procedure is repeated until convergence, i.e., when the covariance norm $\|\Sigma_t\|$ falls below a predefined threshold, or when the maximum number of iterations $T$ is reached. The full algorithm is summarized in Algorithm~\ref{alg:cem}.

\begin{algorithm}[h]
\caption{Cross-Entropy Optimization in Attention Space}
\label{alg:cem}
\begin{algorithmic}[1]
\State \textbf{Input:} Initial mean \(\mu_0\), covariance \(\Sigma_0\), sample size \(m\), elite size \(k\), rollouts \(n\), maximum iterations \(T\)
\For{$t = 0$ to $T-1$}
    \State Sample candidates: \(x_i \sim \mathcal{N}(\mu_t, \Sigma_t), \; i=1,\dots,m\)
    \For{$i=1$ to $m$}
        \State Execute actuator with keypoint \(x_i\) for \(n\) rollouts
        \State Compute average return:
        \[
            R(x_i) = \frac{1}{n} \sum_{j=1}^n r(x_i, j)
        \]
    \EndFor
    \State Select top-$k$ candidates: $\mathcal{S}_t \gets$ indices of largest $R(x_i)$
    \State Update mean:
    \[
        \mu_{t+1} = \frac{1}{k} \sum_{i \in \mathcal{S}_t} x_i
    \]
    \State Update covariance:
    \[
        \Sigma_{t+1} = \frac{1}{k} \sum_{i \in \mathcal{S}_t} (x_i-\mu_{t+1})(x_i-\mu_{t+1})^\top
    \]
    \If{$\|\Sigma_{t+1}\| < \epsilon$}
        \State \textbf{break}
    \EndIf
\EndFor
\State \textbf{Output:} Optimized keypoint distribution \(\mathcal{N}(\mu_t, \Sigma_t)\)
\end{algorithmic}
\end{algorithm}

The reward function employed in our framework is defined based on task completion. For instance, in the cabinet-opening task, the reward is proportional to the degree to which the cabinet door is opened, such that larger opening angles yield higher rewards. To automate the construction of these task-specific reward functions, we leverage a LLM that generates the corresponding reward computation code based on a set of provided examples. We perform all training on an RTX A6000 GPU. For each task step, the combined average training time of RL training and CEM optimization is approximately two hours.

\subsection{Real2sim2real mechanism and real-world deployment}
Since the training process involves simulation, we utilize a 3D generation method \cite{xiang2025structured} to reconstruct real-world scenes into simulations. We then propose a series of measures to overcome the sim2real gap. First, during the RL training phase, we apply domain randomization \cite{tobin2017domain} and introduce small Gaussian noise to the network inputs. Experimental results demonstrate that the policy network remains robust under variations in initial poses, initial velocities, and simulated physical parameters, and can be deployed on the real robot with significantly improved stability. Second, we encode the robot joint angles using their sine and cosine values as network inputs, which effectively reduces discrepancies between simulation and real-world execution. Finally, we incorporate penalties on joint angular velocities and action magnitudes in the reward function, promoting smoother and safer trajectories in real-world scenarios.

Our trained policies can be deployed directly on the real world. The robot is operated at a frequency of 20 Hz. During execution, the mean of the final attention space is used as the target keypoint. For tasks involving object grasping, the gripper remains open until the grasping keypoint is reached, and closes thereafter to perform grasping and manipulation. For non-grasping tasks, the gripper remains closed throughout execution to complete the operation. Importantly, our approach is hardware-free and can be transferred to various robotic platforms, enabling stable execution for complex and long-horizon tasks.

\section{Experiments and Analysis}
In this section, we aim to evaluate whether RoboSeek can precisely extract task-relevant keypoints in real-world scenes and plan safe action executions based on them. We also investigate whether our real2sim2real pipeline can effectively mitigate the sim2real gap in complex, dynamic environments. Moreover, we validate RoboSeek's cross-embodiment generality by testing it on multiple robot platforms.
\begin{figure*}[t]
    \centering
    \includegraphics[width=1\textwidth]{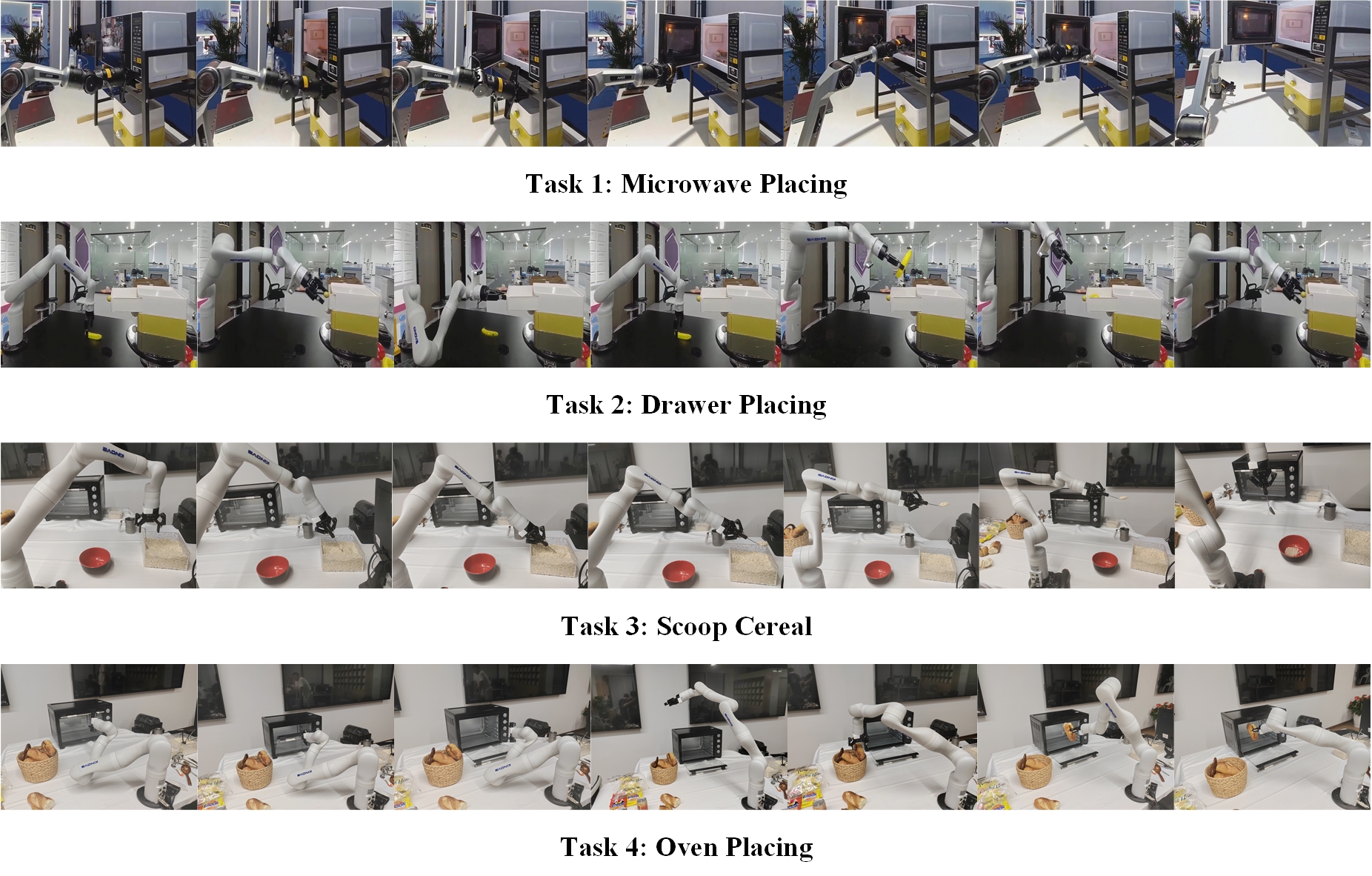}
    \caption{\textbf{Visualization of several complex real-world experiments.} Roboseek demonstrates high success rates and stable control across multi-step tasks, highlighting its capability to learn and operate effectively in real-world physical environments.}
    \label{fig3}
\end{figure*}
\subsection{Experimental setup, task definition, and baselines}
\begin{table}[h]
    \centering
    \begin{tabular}{l c c}
        \toprule
        \textbf{Platform} & \textbf{Task} & \textbf{Success Rate (\%)} \\
        \midrule
        Kinova Gen3  & Barbecue       & 70 \\
                     & Drawer Placing & 80 \\
                     & Oven Placing   & 85 \\
                     & Pour Milk      & 90 \\
                     & Scoop Cereal   & 75 \\
                     & Pour Food      & 60 \\
        \midrule
        Piper        & Drawer Placing    & 85 \\
                     & Microwave Placing & 85 \\
        \bottomrule
    \end{tabular}
    \caption{Success rates of our method on real-robot domestic tasks. Each task is run for 20 trials.}
    \label{tab:success_rates}
\end{table}
We conduct experiments on a Kinova Gen3 and an Agilex Piper robot platform, using an Intel RealSense camera for visual perception. We employ Embodied-R1 \cite{yuan2025embodied} as a high-level perception model to predict keypoints and use SLAT \cite{xiang2025structured} for 3D reconstruction in simulation.By combining depth data from the camera with RGB images, we back-project 2D initial semantic keypoints to obtain their corresponding 3D poses.

We design a suite of long-horizon, complex tasks focused on domestic scenarios. Visualization of several complex tasks is in Figure~\ref{fig3}. On the Kinova Gen3 we evaluate six tasks:
\begin{itemize}[]
    \item \textbf{Barbecue:} Brush oil on the skewers with a brush.
    \item \textbf{Drawer Placing:} Open the drawer, place a banana and close it.
    \item \textbf{Oven Placing:} Open the oven, place a bread into it.
    \item \textbf{Pour Milk:} Pour milk into a bowl.
    \item \textbf{Scoop Cereal:} Scoop cereal into a bowl using a spoon.
    \item \textbf{Pour food:} Lift the lid of the pan and pour the food out.
\end{itemize}
On the Piper platform we evaluate two tasks:
\begin{itemize}[]
    \item \textbf{Drawer Placing:} Open the drawer and pick item out.
    \item \textbf{Microwave Placing:} Open the microwave, grab an onigiri inside and place it.
\end{itemize}

For each task in the real-robot experiments we run twenty trials and report success rates. Trial success is determined by task completion. The detailed per-task success rates on both robot platforms are summarized in Table~\ref{tab:success_rates}.

\begin{figure}[t]
    \centering
    \includegraphics[width=0.5\textwidth]{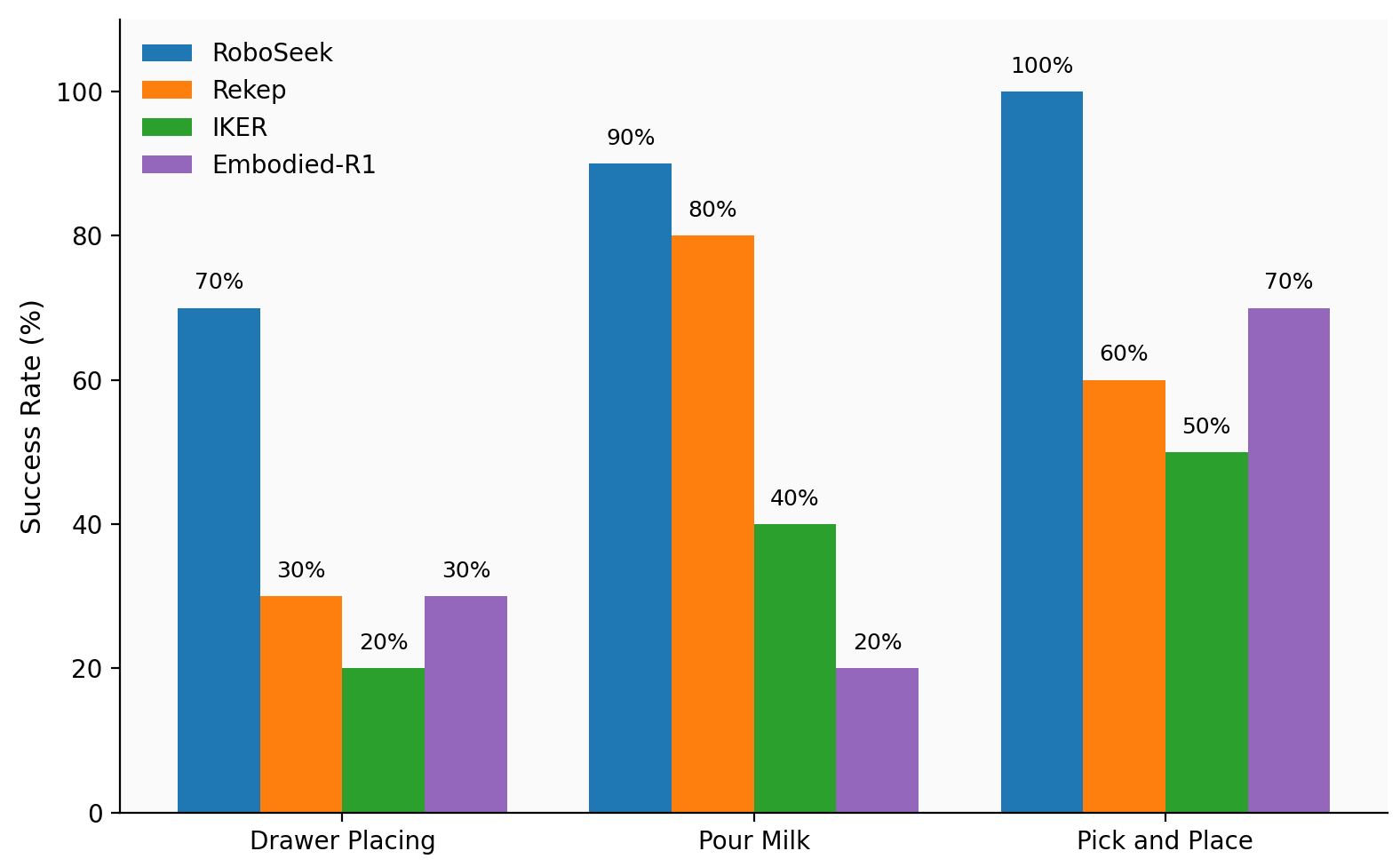}
    \caption{Success rates for different tasks using RoboSeek and Baselines.}
    \label{fig:baseline_comparison}
\end{figure}

As baselines, we compare against Rekep \cite{huang2024rekep}, IKER \cite{patel2025real}, and Embodied-R1 \cite{yuan2025embodied}. Due to environmental limitations of Rekep and IKER, those methods are evaluated only in simulation. For Embodied-R1, we integrate it with a motion planner and evaluate this combination on the real hardware. We select two representative long-horizon tasks and one simple pick-and-place task for a focused comparison between baselines and RoboSeek The comparative results are presented in Figure~\ref{fig:baseline_comparison}. Because the baselines struggle to stably execute long-horizon tasks, we additionally record per-step success counts for baseline methods and aggregate those counts over ten trials to produce the reported success rates.
\subsection{Robust action execution in long-horizon tasks}
Compared to the baselines, RoboSeek shows a clear advantage in executing long-horizon tasks. For example, in the \textit{Pour Milk} task we observed that although Rekep and IKER can generate smooth and temporally coherent motions, they frequently fail to successfully grasp the cup, indicating an inability to extract the task's most relevant keypoints. Embodied-R1, by contrast, is capable of selecting relevant keypoints but lacks a mechanism for predicting keypoint poses, and the classical planner paired with it fails to compute the most effective execution trajectories.

RoboSeek addresses these limitations: its closed-loop corrective behavior enables extraction of task-relevant keypoints, and its embodied executor uses those keypoints to perform effective low-level motions. Consequently, RoboSeek can be stably deployed on complex long-horizon tasks where the baseline methods fail. We find that most of RoboSeek's failures are due to incorrect learning of real-world physical parameters. For instance, in the \textit{Scoop Cereal} task, control commands produced motions that were too fast or applied excessive torque, causing task failure. We believe these issues can be mitigated by improving the fidelity of real-world physics in simulation and further refining the simulation component of our pipeline.

\subsection{Sim2real transfer ability}
We next evaluate the effectiveness of our sim2real transfer mechanisms. We report results from (i) pure simulation, (ii) real-robot deployment without domain randomization, and (iii) real-robot deployment when using only joint-angle inputs as an ablation. Table~\ref{tab:sim2real_comparison} shows our detailed success rates. Although success rates in simulation are marginally higher than on the real robot, the gap is small and does not materially impair deployment performance. These results suggest that our approach substantially reduces the sim2real gap and supports stable real-world deployment. Domain randomization introduces variability during training, making the model more robust to uncertainties encountered in real-world scenarios. Meanwhile, trigonometric encoding provides a more precise representation of the robot's joint angles, mitigating discrepancies between simulation and real-world environments. Consequently, the combination of these techniques is instrumental in achieving effective sim2real transfer.
\begin{table}[h]
    \centering
    \begin{tabular}{l c c c c}
        \toprule
        \textbf{Task} & \textbf{Simulation} & \textbf{Real-World} & \textbf{w/o DR} & \textbf{w/o TE} \\
        \midrule
        Barbecue       & 90\% & 70\% & 20\% & 30\% \\
        Pour Milk      & 95\% & 90\% & 50\% & 45\% \\
        Scoop Cereal   & 90\% & 75\% & 40\% & 40\% \\
        Pour Food      & 80\% & 60\% & 20\% & 35\% \\
        \bottomrule
    \end{tabular}
    \caption{Evaluation of sim2real transfer ability. Success rates are reported for different tasks under several experimental settings.}
    \label{tab:sim2real_comparison}
\end{table}

\section{Conclusion and Limitations}

In this work, we present RoboSeek, an automatic closed-loop pipeline designed for complex manipulation tasks. RoboSeek learns task-relevant semantic keypoints and executes low-level actions through trial-and-error exploration in simulation. Our approach effectively bridges high-level perception models with low-level actuators, enabling continuous learning and optimization of both components. We propose a series of measures to mitigate the sim2real gap, and validate our method across a wide range of long-horizon real-world tasks. Furthermore, RoboSeek demonstrates cross-embodiment capabilities, allowing deployment on multiple robotic platforms. This work provides a new perspective toward achieving general-purpose robotic manipulation in open-world settings and highlights the potential of reinforcement learning for robust low-level action control.

Despite these advances, several limitations remain. First, the network used in this work is relatively small compared to large foundation models, which limits generalization in complex environments. Second, the overall pipeline is time-consuming and tedious, preventing a fast end-to-end response. Finally, the simulation environment cannot perfectly capture complex physical interactions, which restricts the ability to fully address general manipulation tasks.




\bibliographystyle{IEEEtran} 
\bibliography{references}    

\end{CJK*}

\end{document}